\pdfoutput=1
\documentclass[11pt]{article}

\usepackage[final]{coling}

\usepackage{times}
\usepackage{latexsym}

\usepackage[T1]{fontenc}
\usepackage[utf8]{inputenc}

\usepackage{microtype}

\usepackage{inconsolata}

\usepackage{graphicx}

\title{Dynamics of Instruction Fine-Tuning for Chinese Large Language Models}

\author{%
\centerline{Chiyu Song$^{1,2}$ ~~ Zhanchao Zhou$^{1,2}$ ~~ Jianhao Yan$^{1,2}$ } \\ \centerline{\textbf{Yuejiao Fei$^{1,2}$ ~~ Zhenzhong Lan$^{2\text{,}}$\thanks{\ Corresponding authors.} ~~ Yue Zhang$^{2,3,*}$}} \\
\centerline{\normalfont{$^1$Zhejiang University} \quad \quad \normalfont{$^2$School of Engineering, Westlake University}} \\
\centerline{\normalfont{$^3$ Institute of Advanced Technology, Westlake Institute for Advanced Study}}\\
\centerline{\texttt{songchiyu@westlake.edu.cn}}
}

\usepackage{microtype}
\usepackage{color}
\usepackage{multirow}
\usepackage{graphicx}
\usepackage{booktabs}
\usepackage{amsmath}
\usepackage{bbm}
\usepackage{float}
\usepackage{pifont}
\usepackage{colortbl}
\usepackage{tabulary}
\usepackage{etoolbox}
\usepackage{wrapfig}
\usepackage{longtable}

\begin{document}
\maketitle
\begin{abstract}
Instruction tuning is a burgeoning method to elicit the general intelligence of Large Language Models (LLMs). While numerous studies have examined the impact of factors such as data volume and model size on English models, the scaling properties of instruction tuning in other languages remain largely unexplored. In this work, we systematically investigate the effects of data quantity, model size, and data construction methods on instruction tuning for Chinese LLMs. We utilize a newly curated dataset, \textit{DoIT}, which includes over 40,000 high-quality instruction instances covering ten underlying abilities, such as creative writing, code generation, and logical reasoning. Our experiments, conducted on models ranging from 7b to 33b parameters, yield three key findings: (i) While these factors directly affect overall model performance, some abilities are more responsive to scaling, whereas others demonstrate significant resistance. (ii) The scaling sensitivity of different abilities to these factors can be explained by two features: Complexity and Transference. (iii) By tailoring training strategies to their varying sensitivities, specific abilities can be efficiently learned, enhancing performance on two public benchmarks.
\end{abstract}

\section{Introduction}

Large Language Models (LLMs) have shown impressive capabilities across diverse tasks \citep{GPT-3, LLaMA, PaLM, falcon40b, SC-COT, wei2022emergent, Zhao2021CalibrateBU, wei2023zero, Ivison2022HINTHI, Zhang2023MultiTaskIT, radford2019language}, demonstrating their potential for artificial general intelligence \citep{AGI}. A key contributor to this success is instruction tuning, a process involving supervised fine-tuning of LLMs on instruction-output pairs \citep{InstructGPT, Alpaca, vicuna, OPT-IML, LIMA}.

Despite the recognition that various factors, such as data quantity, distribution, and construction methods, directly impact the performance of instruction tuning \citep{LLMsurvey, zhang2023instruction, Tulu}, there remains an inconsistent understanding of their specific roles in shaping model capabilities. For instance, while some studies \citep{FLAN, P3} argue that scaling data volume is crucial for the success of certain tasks, other results \citep{LIMA} suggest that a limited number of instructions can be sufficient. These discrepancies highlight the complexity of instruction tuning and raise concerns about the generalizability of these conclusions to other languages. Moreover, existing studies predominantly focus on English datasets \citep{hestness2017deep, zhang2024scaling}, with a notable lack of comparative analyses in other languages, such as Chinese.

To address the above issues, we introduce \textit{DoIT}, a new Chinese dataset encompassing over 40,000 human-curated instruction instances that span ten distinct LLM abilities. Each data instance is rigorously revised by human annotators to ensure high-quality text and is categorized according to its specific ability. Based on this dataset, we are able to disentangle the effects of each factor by maintaining control over others, thereby providing a clearer understanding of how data volume, parameter size, and construction methods individually influence the development of different abilities. To achieve this, we also employ pre-trained Chinese models such as Chinese-LLaMA \citep{chinese-llama}, Baichuan2 \citep{baichuan2}, and Qwen1.5 \citep{qwen}, resulting in a comprehensive set of instruction-tuned models with sizes ranging from 7 billion to 33 billion parameters.

The results reveal three primary findings:

\begin{enumerate}
    \item Data quantity and parameter size significantly influence overall performance, but each ability develops at different paces during instruction tuning. Abilities such as Creative Writing are more responsive to these factors and can be well-trained with a small amount of data. In contrast, abilities like Ethics show resistance to these changes.

    \item We investigate the reasons behind these discrepancies and identify two features, Complexity and Transference, which can be calculated in low-resource scenarios. These features indicate the potential for ability growth when scaling up data or model parameters.

    \item By leveraging the different sensitivities of abilities to scaling, we can make models learn specific abilities more efficiently and achieve practical performance improvements on two comprehensive benchmarks, CMMLU \citep{cmmlu} and AGIEval \citep{agieval}.
\end{enumerate}

We open-source our codebase, dataset, and model checkpoints for reproducibility and future research\footnote{https://github.com/ChiyuSONG/dynamics-of-instruction-tuning}.

\section{Related Work}
\label{sec:related}
% \vspace{-1mm}

The influence of data factors on instruction tuning has been widely studied, but some conclusions remain inconsistent. Some studies \citep{FLAN, FLAN-T5, zhang2024scaling} suggest that larger datasets improve model performance, while others \citep{LIMA} indicate that a smaller, high-quality dataset can suffice. Although there is evidence that instruction-tuned models generalize well \citep{P3, FLAN}, some argue that this generalization is limited to tasks heavily supported in the training data \citep{falsePromise}. Synthetic data has shown promise \citep{Self-Instruct, yin2023dynosaur}, but the model's capability is constrained by imitating proprietary systems \citep{falsePromise}.

These discrepancies have motivated us to explore how various abilities develop during instruction tuning within a Chinese context. Our research, detailed in Section \ref{sec:study}, reveals significant disparities in the impact of different factors on various abilities. This insight may help reconcile the differing conclusions drawn from prior studies.

Instruction datasets are also crucial for the efficacy of instruction-tuned LLMs, and their construction methods can be broadly categorized into three types: Task-formatted datasets \citep{P3, XP3, FLAN, FLAN-T5, NaturalInstructions, SuperNaturalInstructions} incorporate instances from diverse NLP tasks using human-crafted templates to enable multi-task training. While platforms like PromptSource\cite{PromptSource} have been developed to expand these datasets, concerns about their alignment with real user requests \citep{InstructGPT, LLMsurvey} have led to the exploration of alternative methods. Human-curated datasets \citep{InstructGPT, LIMA, Dolly, openassistant} address the issue above using real-life tasks with human labeling, such as genuine user queries or examination questions. Proprietary models like ChatGPT \citep{ChatGPT} and GPT-4 \citep{GPT-4} employ this data source for training. Unfortunately, these datasets are often not publicly available due to the high cost and effort required. Synthetic datasets \citep{unnaturalInstructions, wizardlm, baize} offer a cost-effective solution by semi-automating instruction generation. One approach is collecting user chats with proprietary models as in ShareGPT\footnote{https://sharegpt.com/}. Self-Instruct \citep{Self-Instruct} is another representative approach, which bootstraps datasets from a small set of seed tasks. This approach has inspired open-source projects like Alpaca \citep{Alpaca} and Vicuna \cite{vicuna}.

\section{DoIT: A New Instruction Dataset}
\label{sec:datasets}

To systematically investigate the roles of data quantity, parameter size, and data construction methods in shaping a range of model abilities, it is necessary to rule out the influence of data quality and establish a controllable data distribution among different abilities. To fulfill these research needs, we introduce \textit{DoIT}, a new human-curated dataset. This dataset contains over 40,000 quality-controlled Chinese instances, categorized into ten distinct ability classes, allowing for tailored experimental setups.

\begin{table}[t]
% \vspace{-10mm}
\centering
\Huge
\resizebox{1\linewidth}{!}{%
\renewcommand{\arraystretch}{1.5}
\begin{tabular}{@{}llllcc@{}}
\toprule
\multirow{2}{*}{Ability} &  & \multirow{2}{*}{Data Source} &  & \multicolumn{2}{c}{Data Size} \\ \cmidrule(l){5-6} 
                       &  &                                               &  & 1st Round & 2nd Round \\ \midrule
STEM - Biology         &  & COIG - Exam \citep{Zhang2023ChineseOI}        &  & 1,200           & 1,242           \\
Humanity - History     &  & COIG - Exam \citep{Zhang2023ChineseOI}        &  & 1,200           & 2,093           \\
Code Generation        &  & Leetcode.cn                                   &  & 1,200           & 5,168           \\
Creative Writing       &  & User Queries from In-House Data               &  & 1,200           & 1,200           \\
Chinese                &  & COIG - Exam \citep{Zhang2023ChineseOI}        &  & 1,200           & 1,650           \\
Dialogue Understanding &  & C3-D \citep{sun2020investigating}             &  & 1,200           & 5,085           \\
Role-play Chat         &  & BELLE \citep{BELLE}                           &  & 1,200           & 1,200           \\
Logical Reasoning      &  & LogiQA2.0 \citep{liu2023logiqa}               &  & 1,200           & 12,951          \\
COT for Grad-Math      &  & PRM800K \citep{lightman2023let}               &  & 1,200           & 11,701           \\
Ethics                 &  & COIG - Human Value \citep{Zhang2023ChineseOI} &  & 1,200           & 1,200           \\ \bottomrule
\end{tabular}
}
\vspace{-1mm}
\caption{The data sources and data size after two rounds of human annotation for each ability category.}
\label{tab:sources}
\vspace{-3mm}
\end{table}

Following the literature reviewed in Section \ref{sec:related}, our human-curated data are derived from real-life contexts, such as academic examinations, online platforms, and user queries. By referencing existing taxonomies \citep{chang2023survey, huang2023ceval}, we focus on practical problems in diverse fields, including ethics, education, engineering, and creative generation, extending beyond basic language capabilities. The ten selected representative abilities encompass a broad spectrum of assessment: (1) STEM subject - Biology, (2) Humanities subject - History, (3) Code Generation, (4) Creative Writing, (5) Language Proficiency - Chinese, (6) Dialogue Understanding, (7) Role-play Chat, (8) Logical Reasoning, (9) Chain of Thought, and (10) Ethics.

To maintain consistent quality across all instances, we employ a three-stage annotation process:

\begin{enumerate}
    
    \item \textbf{Standardization}: Data from diverse sources significantly differ in format, including raw web pages, exam papers, user inputs, and data pre-cleaned by other researchers to different extents. In this stage, we convert them into consistent instruction-output pairs, applying tailored rules for each category to extract relevant text and eliminate duplicates. Notably, the "Chain of Thought" data originated from PRM800K \citep{lightman2023let} is the only non-Chinese source and is translated using the ChatGPT \citep{ChatGPT} API before human review.

    \item \textbf{Human Filtering}: Each item is then reviewed by two independent annotators. They are required to (i) Check the correctness of the text. (ii) Control the diversity of instructions, such as filtering out high-frequency personas in Role-play Chat. (iii) Avoid potential ethical issues in the output, such as biased opinions in Creative Writing. Only items approved by both annotators are accepted, with pass rates ranging from 22.8\% to 98.3\% across different categories and an inter-annotator agreement (IAA) of 0.77.

    \item \textbf{Human Revision}: To ensure adequate data for underrepresented or low-approval categories, we conduct human revision to ensure sufficient numbers for experiments. In this stage, each question is revised or answered by an annotator. Then the answer undergoes the same process as in stage 2, with two additional reviewers determining its validity.

\end{enumerate}

All the hired annotators are native Chinese speakers, hold a bachelor's degree or higher, and dedicate over 1,000 labor hours to annotation. To meet the experimental requirements in Section \ref{sec:study}, the first round of annotation produces 1,000 training data, 100 validation data, and 100 test data for each ability. We then expand the training set to 40k to compare different construction strategies in Section \ref{sec:guidance}. The data sources and sizes for each ability category are outlined in Table \ref{tab:sources}, with examples provided in Appendix \ref{sec:ability examples}.

\section{Experiments}
\label{sec:study}

Employing the human-curated dataset proposed in Section \ref{sec:datasets}, we study the abilities' development in response to alterations in data volume, parameter size, and construction methods. Experiments are conducted under both in-domain and out-of-domain conditions. This section outlines the process of model training, evaluation, and results analysis.

\subsection{Experiment Setup}
\label{sec:epochs}

For quantity-based experiments, we uniformly sample data $d_i$ of size $n$ from each ability $a_i$ within the ten categories $A = \{a_{1}, a_{2}, ..., a_{10}\}$ in our training set. The samples, combined as $D = \bigcup_{i=1}^{10} \{d_i\}$, are utilized for each model training.  We increment the sample size from $n=1$ logarithmically (base 4) to $n=1000$ (totaling 10k instances). Regarding parameter sizes, we train models across a full range of 7b, 13b, and 33b scales. Each training session spans at least 15 epochs, with the corresponding checkpoint saved for evaluation after each epoch. We also compare our human-curated dataset, \textit{DoIT}, with a synthetic dataset proposed by \citet{GPT-4data} for instruction tuning. The synthetic dataset utilizes the Alpaca \citep{Alpaca} instruction pool, created through the Self-Instruct \citep{Self-Instruct} framework, with responses generated by GPT-4 \citep{GPT-4}. By leveraging the cost-effectiveness of synthetic data to acquire a large and diverse set of instances, we can expand our experimental data volume to 41k on this dataset.

Taking into account all these factors, our study requires nearly 500 model checkpoints to draw systematic conclusions. To ensure the generalizability of our findings, we first analyze the scaling properties of different capabilities using the Chinese-LLaMA model \citep{chinese-llama}, which maintains the straightforward architecture of LLaMA \citep{LLaMA} without any modifications. Subsequently, we employ more sophisticated foundation models such as Qwen1.5 \citep{qwen} and Baichuan2 \citep{baichuan2} to further validate how our insights can enhance model training. Detailed hyperparameter choices and training procedures are provided in Appendix \ref{hyper}.

\subsection{Evaluation}

Selecting the optimal checkpoint for instruction-tuning is non-trivial. Prior studies \citep{InstructGPT, LIMA} note that training for more epochs can enhance the model's capabilities despite the risk of overfitting, and usually employ humans for evaluation. In contrast, automated evaluation is a more scalable solution but has long-lasting concerns about reliability in both statistical \citep{bleu, rouge, meteor} and LLM-based \citep{luo2023chatgpt, shen2023large, chiang2023can} metrics. Therefore, to efficiently and accurately scale the evaluation across hundreds of checkpoints, we employ a semi-automated approach to reduce the burden on human annotators.

There are two types of questions in our dataset that correspond to distinct evaluation approaches:

% \vspace{-2mm}

\begin{itemize}
    \item Exact-match questions (e.g., multiple-choice, true/false, fill-in-the-blank) have one exclusive gold answer. Similar to other public benchmarks \citep{mmlu, cmmlu, huang2023ceval, agieval},  we automatically compute the accuracy by comparing generated answers to the ground truth.

    \item Open-ended questions, common in creative writing, role-play chat, and code generation abilities, lack standard answers. We introduce a semi-automated \textit{comparison with distractors} method for these. This method creates distractors (examples shown in Appendix \ref{fig:creative_writing_fine_course} and \ref{fig:code_fine_course}) by manually corrupting each ground truth in two ways: \textbf{Fine-grained corruption} subtly alter some numbers, operators, and terminologies to test the models' performance in modeling details. \textbf{Coarse-grained corruption} creates a distractor that disregards the given instruction but is textually error-free, testing the model's instruction understanding and adherence. A question scores 1 if the language modeling of ground truth $g$ given the instruction $i$ has a lower perplexity (PPL) than any distractor $d_j$, otherwise 0:

\vspace{-7mm}

\begin{gather}
\hspace{-15mm}\small 
\nonumber PPL(g|i)= e^{- \sum_{t=1}^T \log p\left(g_t \mid i, g_{<t}\right)} \\
\hspace{20mm}
\nonumber\text{\small{, $t$ denotes the time series of tokens}}
\end{gather}

\vspace{-8mm}

\begin{gather}
\small\nonumber
Score = \begin{cases} 
1, & \text{if } \min_{j}(PPL(d_j|i)) > PPL(g|i) \\
0, & \text{otherwise}
\end{cases}
\end{gather}

\end{itemize}
\vspace{-2mm}

As outlined in Sections \ref{sec:datasets} and \ref{sec:epochs}, we train 15 checkpoints for each factor setting and reserve 100 instances in both validation and test sets for evaluation. We select the highest-scoring checkpoint after 5 epochs using the validation set and then demonstrate its performance on the test set. Our observations and analysis are discussed in the next subsection.

\begin{figure}[t]
\vspace{-4mm}
  \centering
  \includegraphics[width=0.68\linewidth]{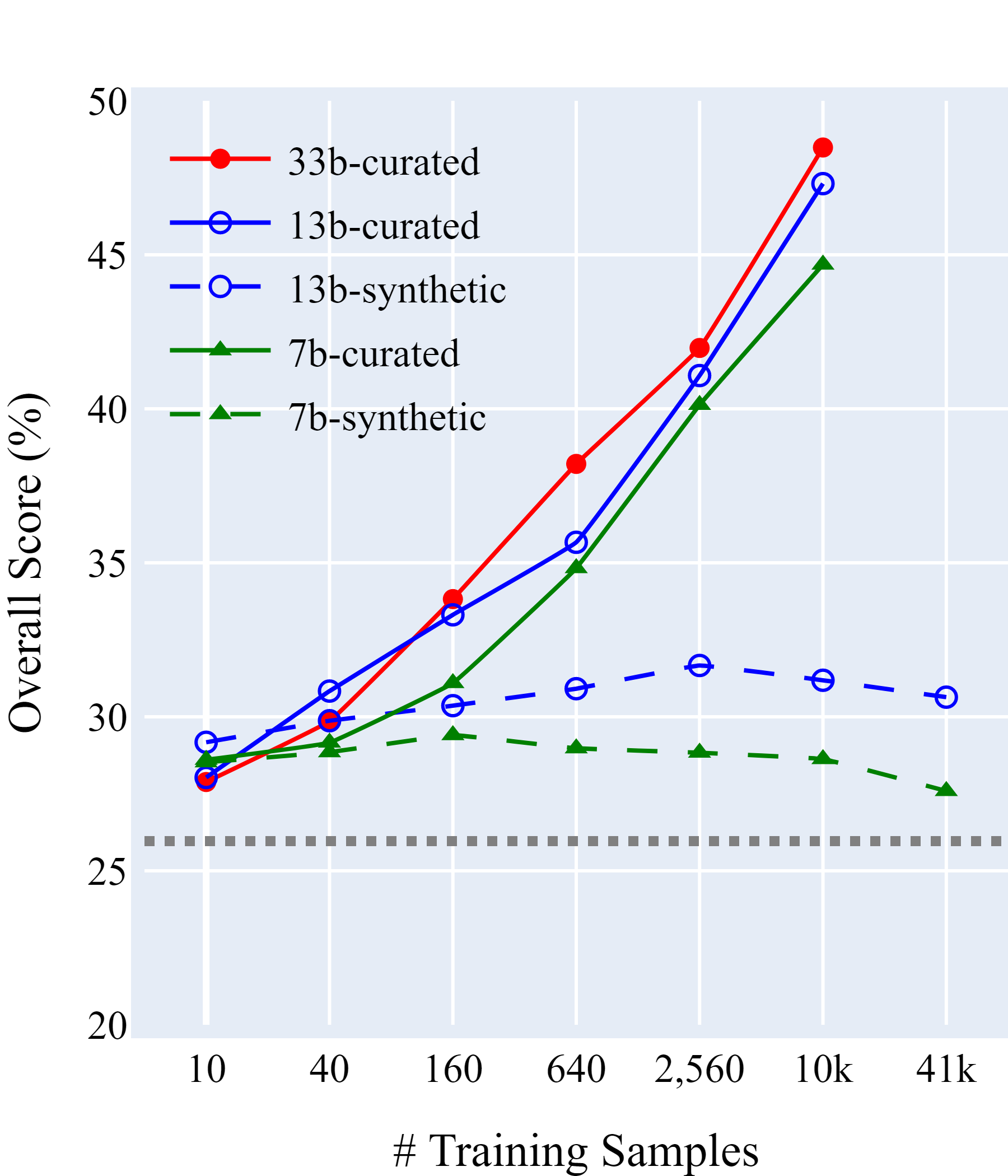}

  \caption{The impact of data volume, parameter scale, and construction method on the overall performance.}

  \label{fig:average}
\vspace{-3mm}
\end{figure}

\begin{figure*}[t]
% \vspace{-7mm}
  \centering
  \includegraphics[width=0.75\linewidth]{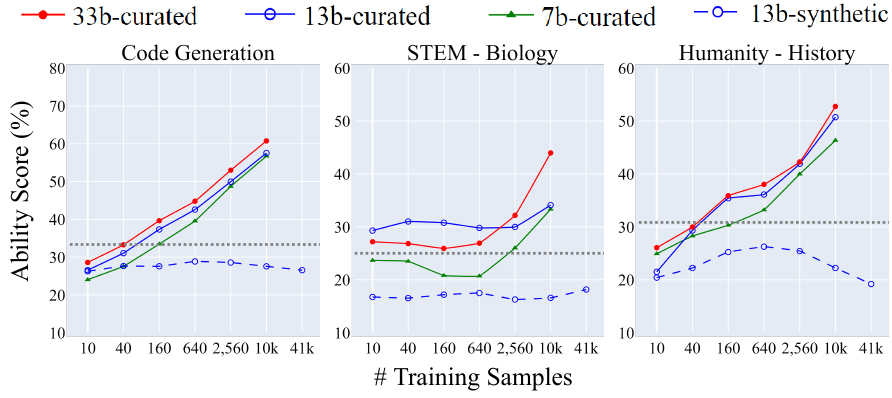}

  \caption{Abilities that are responsive to the data quantity and parameter scale in the human-curated dataset, also comparing the data efficiency of different construction methods with synthetic data.}

  \label{fig:responsive}

\end{figure*}

\begin{figure*}[t]
  \centering
  \includegraphics[width=1\linewidth]{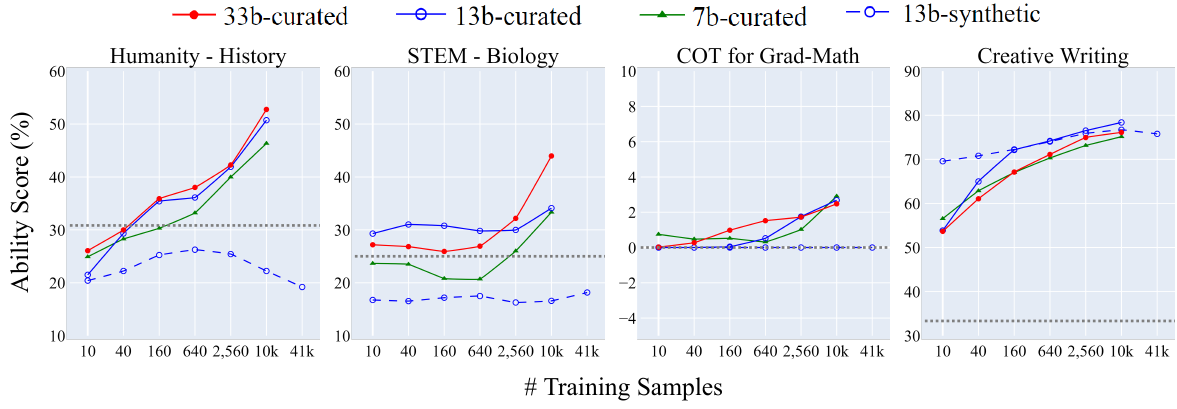}

  \caption{Comparison of abilities with varying sensitivities to data scaling in the human-curated dataset, also comparing the data efficiency of different construction methods with synthetic data.}

  \label{fig:data-scaling}
% \vspace{-2mm}
\end{figure*}

\subsection{Results and Analysis}
\label{sec:analysis}

We analyze the effect of data volume, parameter size, and construction method. Their impact on overall model performance is illustrated in Fig \ref{fig:average}, where the x-axis represents changes in data volume and the y-axis represents the average scores across ten in-domain evaluations plus three out-of-domain abilities. Lines of different colors and symbols represent models with different parameter sizes. We also have a grey dotted line representing the score of random guesses. When scaling the number of training instances, there is a substantial discrepancy on the performance of models trained on human-curated data (depicted by solid lines) and synthetic data (depicted by dashed lines).

Moreover, the overall trend is not universally applicable to different abilities when we observe them in the next section. Subsequently, we quantify the scaling sensitivity of each ability by investigating the relationship between its task accuracy and the factors above. We further analyze two interpretable features that potentially forge different scaling sensitivities across these abilities.

\begin{figure*}[t]
% \vspace{-7mm}

  \centering
  \includegraphics[width=1\linewidth]{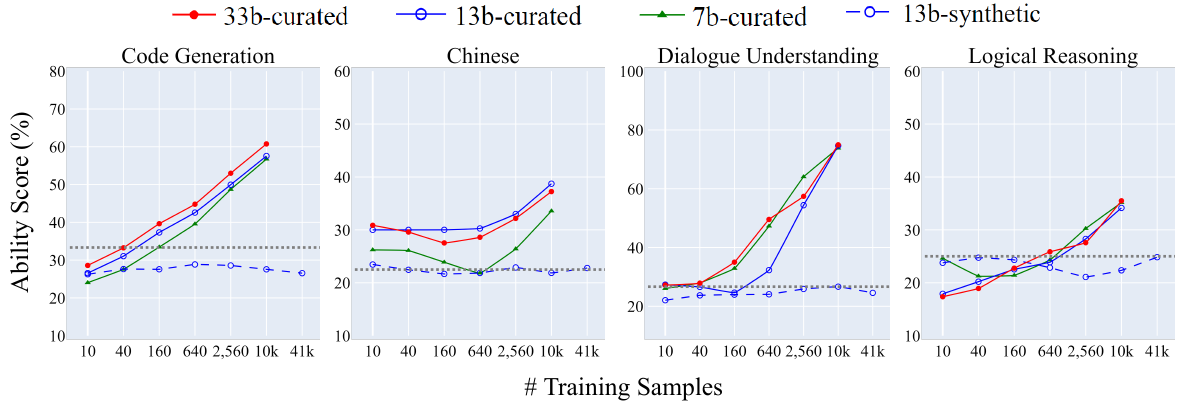}

  \caption{Comparison of abilities with varying sensitivities to parameter size scaling in the human-curated dataset, also comparing the data efficiency of different construction methods with synthetic data.}

  \label{fig:param-scaling}

\end{figure*}

\begin{figure}[t]
\vspace{-3mm}
  \centering
  \includegraphics[width=1.02\linewidth]{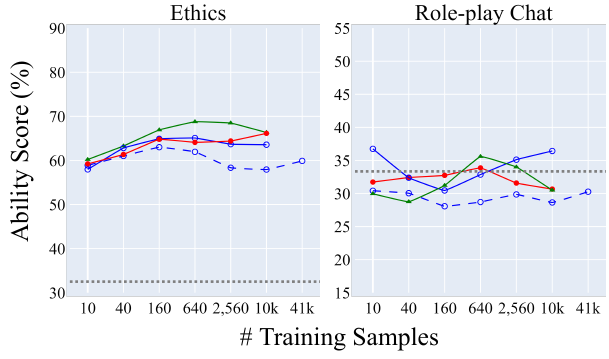}

\vspace{-2mm}
  \caption{Abilities that are resistant to the data quantity and parameter scale in the human-curated dataset, also comparing the data efficiency of different construction methods with synthetic data.}

  \label{fig:resistant}

\end{figure}

\subsubsection{Disparities in ability growth trajectories}
\label{compare}

\begin{figure*}[ht]
% \vspace{-11mm}
  \centering
  \includegraphics[width=0.75\linewidth]{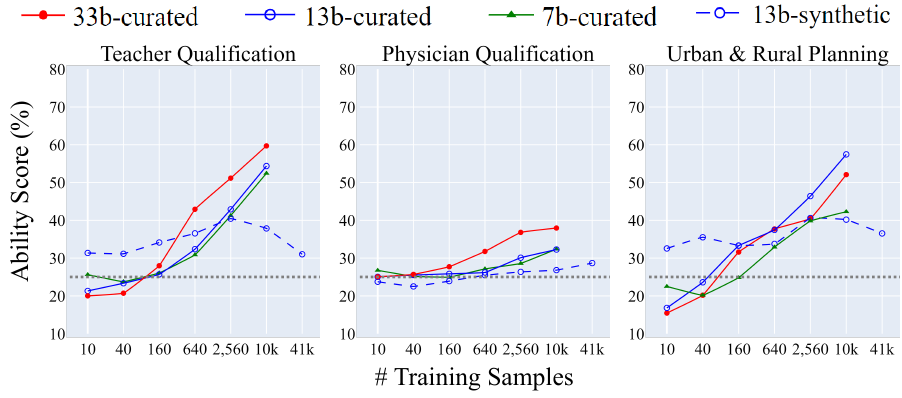}

\vspace{-1mm}
  \caption{Growth paces of out-of-domain abilities that not included in the human-curated dataset, also comparing the data efficiency of different construction methods with synthetic data.}

  \label{fig:ood}

\vspace{-4mm}

\end{figure*}

\begin{figure}[ht]
% \vspace{-2mm}
  \centering
  \includegraphics[width=1.03\linewidth]{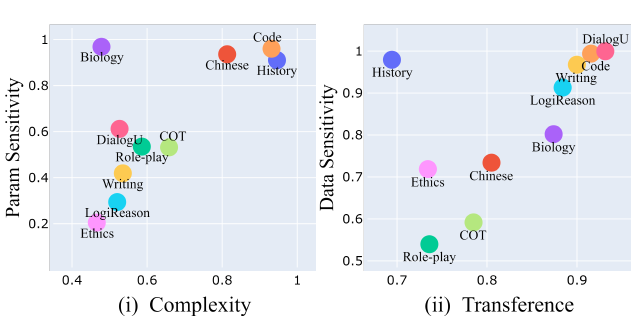}

  \caption{Two interpretable features, (i) Complexity and (ii) Transference, of different abilities demonstrate linear relationships with their sensitivities to scaling in parameter size and data volume. These features can infer the growth of abilities after scale-up, as discussed in Section \ref{sec:guidance}. Sensitivity values are normalized to (0, 1) for visualization.}

  \label{fig:sense}
% \vspace{2mm}
\end{figure}

We present the empirical results for each ability in this section, exhibiting their distinct growth paces when facing factor changes:

    \noindent \textbf{Abilities responsive to scaling}: Some abilities such as Code Generation, STEM-Biology, and Humanity-History are responsive to factor changes. As illustrated in Fig \ref{fig:responsive}, they show clear upward trends with the growth of data volume and parameter scale.

    \noindent \textbf{Varying sensitivities to data scaling}: As depicted in Fig \ref{fig:data-scaling}, the rate of improvement is not uniform across abilities. This figure reveals varying degrees of data scaling sensitivity, with Creative Writing being a notable case. The slope of its curve gradually disappears, indicating a plateau with limited data expansion.

    \noindent \textbf{Varying sensitivities to parameter scaling}: Fig \ref{fig:param-scaling} demonstrates that the impact of parameter size scaling also varies among abilities. From left to right in the figure, their curves for different model sizes become increasingly intertwined, indicating the insensitivity to this change.

    \noindent \textbf{Abilities resistant to scaling}: As shown in Fig \ref{fig:resistant}, certain abilities like Ethics and Role-play Chat appear to resist both factors and maintain stagnant scores across all changes. This lack of progress implies that supervised fine-tuning (SFT) alone may not effectively advance these abilities, warranting the investigation of approaches beyond it, such as reinforcement learning from human feedback (RLHF) \citep{InstructGPT, webGPT}.

    \noindent \textbf{Out-of-domain evaluation}: Beyond in-domain abilities, Figure \ref{fig:ood} shows model performance on three out-of-distribution (OOD) tasks from the C-Eval datasets \citep{huang2023ceval}: Teacher Qualification, Physician Qualification, and Urban and Rural Planning. The observed growth trends suggest robust cross-ability generalization. Similar to in-domain evaluations, these OOD tasks exhibit diverse responses to variations in data quantity and parameter scale.

    \noindent \textbf{Human-curated vs. Synthetic}: Figures \ref{fig:responsive}-\ref{fig:ood} also present the results from models trained on synthetic data generated by GPT-4 \citep{GPT-4data}. We evaluate both 7B and 13B models, which yield analogous conclusions. For simplicity, only the results for the 13B model are plotted, with the 7B results included in Appendix \ref{fig:7b}. Our findings align with previous studies \citep{falsePromise}, indicating that synthetic data is only effective for learning partial abilities. Additionally, Figure \ref{fig:average} demonstrates that increasing the volume of synthetic data does not continuously improve model performance. We further empirically demonstrate in Appendix \ref{mix} that even when synthetic data is combined with human-curated data, its effectiveness still has an upper limit. Consequently, in subsequent experiments, we focus solely on exploring the scaling properties of human-curated data.

% \end{itemize}

\subsubsection{Understanding Diverse Scaling Behaviors}
\label{behavior}
To understand the varying scaling properties of abilities, a notable observation from Section \ref{compare} is that abilities tied to professional (academic) knowledge are more sensitive to parameter scaling. We define such common feature as \textbf{Complexity}, indicating these abilities are inherently \textit{"more challenging for language modeling"} and \textit{"benefit less from the training of other abilities"}. We hypothesize that Complexity is associated with how different abilities respond to model size changes.

To examine the relationship between Complexity and parameter sensitivity, we first quantify the sensitivities of individual abilities. Adopting the scaling law function similar to \citet{kaplan2020scaling}, we model the task score $ACC$ (averaged across varying data volume) as a function of model size $N$ for each ability $i$:

\vspace{-6mm}
\begin{gather}
\label{eq1}
\small
    exp(\text{ACC}_{i}) = (exp(c_{i}) \cdot N)^{\alpha_{i}}
    \text{\small{, $c_{i}$ is constant}}
\end{gather}
\vspace{-5mm}

Here, the exponent $\alpha_{i}$ represents the rate of accuracy improvement with increasing model size, indicating the \textit{scaling sensitivity}.

\vspace{-6mm}
\begin{gather}
\small
    \text{ACC}_{i} = \alpha_{i} \cdot \log(N)+\alpha_{i} \cdot c_{i}
\end{gather}
\vspace{-5mm}

We further demonstrate that Complexity can be measured in a low-resource setting by fine-tuning separate 7b models with only 64 data points per ability. According to its definition, Complexity is calculated as a weighted sum of the test loss ($L$) for each model trained individually on ability $i$, along with its accuracy achieved by training on ability $j$ data, $Acc(j,i)$, improved over the foundation model's performance $Acc(f,i)$:

\vspace{-7mm}
\begin{gather}
\nonumber
\small
   \text{Complexity}_{i} = w_{1} \cdot {\text{L}_{i}} - w_{2} \cdot {\sum_{\text{j} \neq \text{i}} ( \text{Acc}(j, i) - \text{Acc}(f, i) )}
\end{gather}
\vspace{-4mm}

Results depicted in Figure \ref{fig:sense} (i) show a clear linear relationship between Complexity and the sensitivity to parameter scale, indicating that even with minimal resources (7b model size, 64 data points), we can forecast how abilities will develop with increased model size.

Correspondingly, we have also computed the \textbf{Transference} for each ability, which reflects \textit{"how well ability $i$ data enhances other abilities $j$"} via this formula:

\vspace{-7mm}
\begin{gather}
\nonumber
\small
   \text{Transference}_i = w \cdot \sum_{\text{j} \neq \text{i}} ( \text{Acc}(i, j) - \text{Acc}(j, j) )
\end{gather}
\vspace{-4mm}

By substituting model size ($N$) with data volume ($D$) in Equation (\ref{eq1}), we can also evaluate each ability's sensitivity to data volume. Figure \ref{fig:sense} (ii) shows that Transference is linearly related to data scaling sensitivity. This confirms that using checkpoints trained with limited data can also infer the impact of enlarged data volume on ability development.

\section{Guidance on Model Training}
\label{sec:guidance}

After understanding that \textit{"abilities react differently to factor changes"} and learning \textit{"how to estimate the sensitivities of different abilities to scaling,"} we delve deeper into whether these varying sensitivities can enhance the efficiency of learning specific or overall model abilities. At this stage, we employ two more advanced foundation models, Qwen1.5\citep{qwen} and Baichuan2\citep{baichuan2}, to conduct our experiments.

\subsection{Learning  a Specific Ability}
\label{single}

When learning a new ability, due to its unique sensitivity to different factors, the efficiency of improvement can vary significantly depending on how resources are allocated between computational power and data annotation. We select two representative abilities to test the effectiveness of model training according to their distinct sensitivities:

\textbf{Logical Reasoning}: As demonstrated in Figure \ref{fig:sense}, this ability is data-biased, meaning it is more sensitive to data scaling than to parameter scaling.

\textbf{Novel Generation}\footnote{https://huggingface.co/datasets/zxbsmk/webnovel\_cn}: This OOD task involves continuing a novel from a given starting point. According to the calculations described in Section \ref{behavior}, Novel Generation is parameter-biased with a parameter sensitivity of 0.92 and a data sensitivity of 0.54. This indicates that increasing the model's parameter size is more effective for enhancing this ability than increasing the data volume.

Table \ref{tab:specific} illustrates the performance of Qwen1.5 and Baichuan2 when trained with varying amounts of data and parameters, following the evaluation methodology outlined in Section \ref{sec:study}. The results show that for the data-biased ability, Logical Reasoning, increasing the data volume from 2000 to 4000 can yield performance that matches or exceeds that of models with 13/14B parameters. This suggests that annotating more data can effectively reduce computational resource requirements. Conversely, for the parameter-biased ability, Novel Generation, even a fivefold increase in data volume fails to match the performance of larger models. Therefore, for such abilities, increasing the model's parameter size is the more effective strategy.

\begin{table}[t]
% \vspace{-10mm}
\Large
\centering
\resizebox{1\linewidth}{!}{%
\begin{tabular}{@{}llccccccc@{}}
\toprule
 &  & \multirow{2}{*}{\begin{tabular}[c]{@{}c@{}}Parameter\\ Scale\end{tabular}} &  & \multicolumn{2}{c}{Novel Generation} &  & \multicolumn{2}{c}{Logical Reasoning} \\ \cmidrule(lr){5-6} \cmidrule(l){8-9} 
                           &  &     &  & Data Size & Score         &  & Data Size & Score         \\ \midrule
\multirow{3}{*}{Qwen1.5}   &  & 7B  &  & 2,000     & 53.0          &  & 2,000     & 26.0          \\
                           &  & 7B  &  & 10,000     & 52.0 &  & 4,000    & \textbf{37.0}          \\
                           &  & 14B &  & 2,000     & \textbf{62.0}          &  & 2,000     & 34.0 \\ \midrule
\multirow{3}{*}{Baichuan2} &  & 7B  &  & 2,000     &   37.0        &  & 2,000     & 25.0          \\
                           &  & 7B  &  & 10,000     & 38.0 &  & 4,000    & \textbf{35.0}          \\
                           &  & 13B &  & 2,000     & \textbf{57.0} &  & 2,000     & \textbf{36.0} \\ \bottomrule
\end{tabular}
}
\vspace{-2mm}
\caption{Parameter-biased ability (Novel Generation) shows significantly greater performance gains from increasing model parameters compared to adding more data. Conversely, data-biased ability (Logical Reasoning) benefits more from additional labeled data, as discussed in Section \ref{single}.}
\label{tab:specific}
\end{table}

\subsection{Learning Comprehensive Abilities}
\label{recon}

% \vspace{-1mm}

If an ability exhibits low sensitivity to both data and parameter variations, it implies that this ability cannot be effectively developed through instruction tuning. We investigate whether reducing the amount of such insensitive data while increasing the proportion of other, more responsive data can enhance the efficiency of model training.

We evaluate three data-mixing strategies:

\textbf{Baseline}: We use the model trained on 1,000 instances per ability (totaling 10,000 instances) from Section \ref{sec:study} as the baseline.

\textbf{Reconstruct}: Ethics and Role-play Chat, which show low sensitivity to all factor changes in Section \ref{sec:analysis}, are reduced to 64 instances each (corresponding to their relatively higher points in Figure \ref{fig:resistant}). The total number of instances is maintained at 10,000 by uniformly increasing the data volume of other categories. Creative Writing remains at 1,000 instances, as Figure 3 indicates that this is sufficient for it to reach its performance plateau.

\textbf{Maximum}: We further expand the data volume based on the same insights, keeping Ethics and Role-play Chat at 64 instances each and Creative Writing at 1,000 instances. Other abilities are scaled up following the procedures outlined in Section \ref{sec:datasets}, with specific quantities detailed in Table \ref{tab:sources}. This expansion results in an unbalanced dataset due to the varying difficulty in annotating each ability.

For evaluation, we use two comprehensive benchmarks: AGIEval \citep{agieval} and CMMLU \citep{cmmlu}. AGIEval \citep{agieval} assesses the general capabilities of LLMs in tasks related to human cognition and problem-solving, focusing specifically on the multiple-choice questions within its three Chinese subsets. CMMLU \citep{cmmlu}, akin to MMLU \citep{mmlu}, evaluates LLMs' knowledge and reasoning capabilities within a Chinese cultural context, spanning 67 diverse subjects from elementary to advanced professional levels.

Models with 7b parameters are trained for each data-mixing strategy and their performance is tested at epoch 10 on both benchmarks under 0-shot and 5-shot settings. Table \ref{tab:reconstruct} marks the results that show improvement over the baseline with $\uparrow$. In our experiments with three distinct foundation models, both new strategies generally outperform the "Baseline" approach. Specifically, the "Reconstruct" strategy improves AGIEval scores by 1\%-3\% without increasing the data volume. Moreover, the "Maximum" strategy further enhances performance across all evaluated abilities.

\begin{table}[t]
% \vspace{-10mm}
\Huge
\centering
\resizebox{1.03\linewidth}{!}{%
\begin{tabular}{@{}lccccclll@{}}
\toprule
\multirow{2}{*}{Models} & \multirow{2}{*}{Data Quantity} & \multicolumn{3}{c}{AGIEval - 0shot} & \multirow{2}{*}{} & \multicolumn{3}{c}{CMMLU - 0shot} \\ \cmidrule(lr){3-5} \cmidrule(l){7-9} 
 &
   &
  CLLaMA &
  Baichuan2 &
  Qwen1.5 &
   &
  \multicolumn{1}{c}{CLLaMA} &
  \multicolumn{1}{c}{Baichuan2} &
  \multicolumn{1}{c}{Qwen1.5} \\ \midrule
Baseline &
  10k &
  34.64 &
  42.15 &
  69.08 &
   &
  36.75 &
  52.60 &
  72.71 \\
Reconstruct &
  10k &
  35.43$\uparrow$ &
  45.09 &
  69.56$\uparrow$ &
   &
  36.85$\uparrow$ &
  53.00$\uparrow$ &
  73.07$\uparrow$ \\
Maximum &
  40k &
  37.61$\uparrow$ &
  46.59$\uparrow$ &
  69.21$\uparrow$ &
   &
  37.28$\uparrow$ &
  56.50$\uparrow$ &
  72.33 \\ \midrule
\multirow{2}{*}{} &
  \multirow{2}{*}{} &
  \multicolumn{3}{c}{AGIEval - 5shot} &
  \multirow{2}{*}{} &
  \multicolumn{3}{c}{CMMLU - 5shot} \\ \cmidrule(lr){3-5} \cmidrule(l){7-9} 
 &
   &
  CLLaMA &
  Baichuan2 &
  Qwen1.5 &
   &
  \multicolumn{1}{c}{CLLaMA} &
  \multicolumn{1}{c}{Baichuan2} &
  \multicolumn{1}{c}{Qwen1.5} \\ \midrule
Baseline &
  10k &
  31.01 &
  47.03 &
  70.12 &
   &
  35.14 &
  54.87 &
  71.97 \\
Reconstruct &
  10k &
  32.37$\uparrow$ &
  48.46$\uparrow$ &
  70.97$\uparrow$ &
   &
  35.89$\uparrow$ &
  55.00$\uparrow$ &
  71.82 \\
Maximum &
  40k &
  33.57$\uparrow$ &
  53.12$\uparrow$ &
  70.96$\uparrow$ &
   &
  37.16$\uparrow$ &
  58.02$\uparrow$ &
  71.18 \\ \bottomrule
\end{tabular}
}
% \vspace{-2mm}
\caption{Comparing the performance of three data-mixing strategies on two benchmarks, evaluated using checkpoints at epoch 10 with a parameter size of 7b. Scores superior to the baseline are marked with $\uparrow$.}
\label{tab:reconstruct}
\end{table}

\section{Conclusion}
This research introduces a novel, human-curated Chinese dataset comprising over 40,000 instruction instances across ten ability categories. This dataset facilitates the investigation of how the growth of large language model (LLM) abilities is influenced by data volume, parameter size, and data construction methods during instruction tuning. We are the first to disentangle the effects of these factors by examining a comprehensive set of over 500 model checkpoints, ranging from 7 billion to 33 billion parameters. Our findings reveal that the impact of increasing data volume and model size varies across different abilities. We identify two key features, Complexity and Transference, which can predict ability growth in low-resource scenarios. Guided by these findings, we enhance the effectiveness of learning specific tasks and achieve better comprehensive abilities on benchmarks such as CMMLU and AGIEval. We provide open access to our codebase, dataset, and model checkpoints to enhance reproducibility and contribute to future research.

\section*{Acknowledgements}
This publication has been supported by the Pioneer and "Leading Goose" R\&D Program of Zhejiang under Grant Number 2022SDXHDX0003, and the project titled "Research on Key Technologies of Large-scale Models and Their Application in Education" under Grant Number 10311003H042401.

% \vspace{-2mm}

\bibliography{custom}

\appendix
\section{Appendix}

\begin{figure}[ht]
% \vspace{-2mm}
  \centering
  \includegraphics[width=1.03\linewidth]{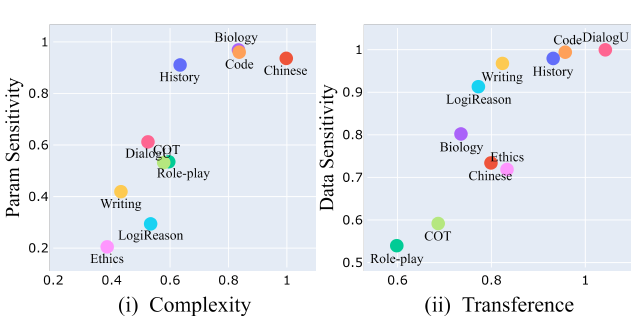}

  \caption{Two interpretable features, (i) Complexity and (ii) Transference, of different abilities demonstrate linear relationships with their sensitivities to scaling in parameter size and data volume. These features are calculated using 1000 data points instead of 64 in Section \ref{sec:guidance}. Sensitivity values are normalized to (0, 1) for visualization.}

  \label{fig:sense-1000}
% \vspace{2mm}
\end{figure}

\subsection{Training Procedures}
\label{hyper}

Our hyperparameter choices are generally in line with \citet{LIMA}, using AdamW optimizer with $\beta_{1}=0.9$, $\beta_{2}=0.95$ and $weight\_decay=0.1$. The peak learning rate is set to $1e$-$5$ and linearly decays to $1e$-$6$ by the end of training. The batch size is 64, with inputs exceeding 2048 tokens trimmed. Referring to the protocol in \citet{song2023uni}, we also incorporate two speaker tokens, $<\text{user}>$ and $<\text{assistant}>$, to segment the utterances from instruction and output for training and inference. 

\newpage

\subsection{Examples of each ability category}
\label{sec:ability examples}

\onecolumn
\begin{longtable*}[p]{p{.2\textwidth} p{.75\textwidth}}
\toprule
    Ability &                                                            Example 
    \\ \midrule
\midrule
\endfirsthead

\toprule
    Ability &                                                            Example 
    \\ \midrule
\midrule
\endhead
\midrule
\multicolumn{2}{r}{{Continued on next page}} \\ \midrule
\midrule
\endfoot
\bottomrule
\endlastfoot
    Creative writing &       
    \textbf{Prompt:}
    {\newline}
    Write a copywriting for the 'Wangshan Hawthorn'.
    {\newline}
    {\newline}
    \textbf{Response:}
    {\newline}
    Folded hopes' branches,{\newline}
    Lightly swaying within hawthorn's embrace from within.{\newline}
    
    A gentle touch on the tongue,{\newline}
    Tartness tracing the intricate pathways of taste buds,{\newline}
    Unfolding and spreading deep into the recesses of lips and teeth.{\newline}
    
    Having savored the bittersweet fruit,{\newline}
    Have you ever witnessed the bloom of its origin?{\newline}
    
    Remember every tangy journey,{\newline}
    Never forget those times when the flowers were in full bloom.{\newline}
    
    \#GoodHopeWater \#Wangshan Hawthorn \#AdvertisingCopy{\newline}
\\ \midrule
    Dialogue \newline Understanding &       
    \textbf{Prompt:}
    {\newline}
    Man: What kind of animal do you like?{\newline}
    
    Woman: Elephants. They have long trunks, big ears, and they look so clumsy, which makes them adorable.{\newline}

    Man: Elephants aren't clumsy, though. They often help humans with work.{\newline}

    Woman: I said they "look" clumsy, you know.{\newline}
    
    Why does the woman like elephants?{\newline}
    A. Elephants are big.{\newline}
    B. Elephants are clumsy.{\newline}
    C. Elephants are adorable.{\newline}
    D. Elephants are smart.{\newline} 
    {\newline}
    \textbf{Response:}
    {\newline}
    C{\newline}
\\ \midrule
    Role-play Chat &       
    \textbf{Prompt:}
    {\newline}
    Ali is the CEO of a young startup company, possessing an innovative mindset and a pursuit of excellence. He is actively seeking professional talent to assist him in realizing his business vision.{\newline}
    John, on the other hand, is an experienced HR manager with profound expertise and extensive experience in talent acquisition.{\newline}
    Generate a short conversation happening between them.
    {\newline}
    {\newline}
    \textbf{Response:}
    {\newline}
    Ali: Hey John, thank you for taking the time to chat with me. I'm really interested in learning more about your career experience, and I'm quite impressed by your expertise and experience in HR.{\newline}
    
    John: Thank you for the invitation, Ali. I'm very interested in your company and its business vision. Can you share your perspective on talent recruitment and development within the company?{\newline}

    Ali: In the early stages of our startup, we're in great need of visionary and passionate talents. We're looking for individuals who can contribute to the company's growth and help us achieve our vision together. We'd like to better understand how to attract, select, and nurture these talents to ensure the ongoing success of our company.{\newline}
    
    John: Your thoughts align well with our priorities. I believe attracting high-caliber talent is a crucial mission for any startup. There are various ways our company can achieve these goals, such as through recruitment, building reputation, and talent development. However, I think the most critical aspect is having a tailored talent management and development plan to ensure that individuals who fit our company's needs can grow within the organization and continually enhance their skills and knowledge.{\newline}
    
    Ali: That sounds excellent. I think we can delve deeper into this project.{\newline}
\\ \midrule
    STEM - Biology &       
    \textbf{Prompt:}
    {\newline}
    Please choose the appropriate option based on the content of the question.{\newline}
    Regarding enzymes, which statement is correct?{\newline}
    A. Enzymes provide the necessary activation energy for the reaction.{\newline}
    B. Changes in enzyme activity are unrelated to changes in the enzyme's environment.{\newline}
    C. Changes in enzyme structure can result in the loss of its active site or its entire activity.{\newline}
    D. Enzyme molecules are immediately degraded into amino acids after catalyzing a reaction.{\newline}
    {\newline}
    {\newline}
    \textbf{Response:}
    {\newline}
    C{\newline}
\\ \midrule
    Chinese &       
    \textbf{Prompt:}
    {\newline}
    Reading the following poem and anwsering the question,{\newline}

    Moon Song of Mount Emei by Li Bai{\newline}
    On Emei Mountain, a crescent moon graces autumn's sky,{\newline}
    Its reflection dances upon the tranquil water of the Pingqiang River nearby.{\newline}
    By night, I set forth on the clear stream towards the Three Gorges afar,{\newline}
    Thoughts of you linger during my journey, for you are not in sight in Yuzhou.{\newline}
    
    Please describe the scene painted by the lines "On Emei Mountain, a crescent moon graces autumn's sky, Its reflection dances upon the tranquil water of the Pingqiang River nearby" in vivid words.{\newline}
    {\newline}
    {\newline}
    \textbf{Response:}
    {\newline}
    Lofty Emei Mountain stands tall and grand, with a crescent autumn moon hanging amidst its peaks, resembling the mountain itself spilling the moonlight. The pristine moonbeam is mirrored on the surface of the tranquil Pingqiang River, which flows gently, carrying the moon's reflection along its course.{\newline}
\\ \midrule
    Ethics &       
    \textbf{Prompt:}
    {\newline}
    Some people admire Western culture and believe that Chinese culture is conservative and outdated. How should you respond to them? Drawing on your knowledge, share your perspective.
    {\newline}
    {\newline}
    \textbf{Response:}
    {\newline}
    1. Different cultures have different values, and it's not appropriate to judge other cultures solely by Western standards.{\newline}
    2. Despite being relatively conservative, Chinese culture possesses unique charm and profound depth.{\newline}
    3. Rather than excluding other cultures, it's important to maintain an open-minded attitude and learn from, as well as incorporate, the beneficial aspects of other cultures.{\newline}
    4. One should avoid unilaterally idealizing any single culture and, instead, choose and integrate various cultures based on practical circumstances.{\newline}
\\ \midrule
    Logical Reasoning &       
    \textbf{Prompt:}
    {\newline}
    DM advertising, also known as direct mail advertising, involves delivering promotional materials to consumers through mailing, gifting, or other means, directly to their homes or places of business.{\newline}

    According to the definition provided, which of the following is not considered as DM advertising?{\newline}
    A. Samples of sunscreen lotion given away at cosmetics counters in department stores.{\newline}
    B. Flyers promoting a commercial real estate project distributed on crowded streets.{\newline}
    C. Coupons for a large chain of fast-food restaurants received in an email inbox.{\newline}
    D. Two screen protectors received as a gift when purchasing a mobile phone at an electronics store.{\newline}
    {\newline}
    {\newline}
    \textbf{Response:}
    {\newline}
    D{\newline}
\\ \midrule
    COT for Grad-Math &       
    \textbf{Prompt:}
    {\newline}
    Solve the inverse matrix of \[\begin{pmatrix} 2 & 3 \\ -1 & 7 \end{pmatrix}\]. Output a zero matrix if the inverse matrix doesn't exists 
    {\newline}
    {\newline}
    \textbf{Response:}
    {\newline}
    To solve the inverse matrix of a $2\times 2$ matrix, we can use the following formula
    \[\begin{pmatrix} a & b \\ c & d \end{pmatrix}^{-1} = \frac{1}{ad-bc}\begin{pmatrix} d & -b \\ -c & a \end{pmatrix},\]
    and the condition holds as long as its determinant  $ad-bc$ is non-zero.
    {\newline}
    
    In this example, $2\times 7 - 3\times (-1) = 14 + 3 = 17$, which is not zero, thus there exsits an inverse matrix.
    {\newline}
    
    Apply the formula, and we will have:
    \[\begin{pmatrix} 2 & 3 \\ -1 & 7 \end{pmatrix}^{-1} = \frac{1}{17}\begin{pmatrix} 7 & -3 \\ 1 & 2 \end{pmatrix}.\]
    {\newline}
    The final anwser is
    \[\frac{1}{17}\begin{pmatrix} 7 & -3 \\ 1 & 2 \end{pmatrix}.\]{\newline}
\\ \midrule
    Humanity - History &       
    \textbf{Prompt:}
    {\newline}
    True or False: The three major battles of Liaoshen, Huaihai, and Crossing the Yangtze River essentially annihilated the main forces of the Nationalist Army, accelerating the overall victory of the Liberation War nationwide. ( ){\newline}
    {\newline}
    {\newline}
    \textbf{Response:}
    {\newline}
    False{\newline}
\\ \midrule
    Code Generation &       
        \begin{minipage}[t]{\linewidth}
        \textbf{Prompt:}
        {\newline}
        You are given 2 integers left and right, indicating interval [left, right], return the bitwise AND results of all numbers within this interval (including the left and right endpoints).{\newline}
     
        Example 1:{\newline}
        Input: left = 5, right = 7{\newline}
        Output: 4{\newline}
        Example 2: {\newline}
        Input: left = 0, right = 0{\newline}
        Output: 0{\newline}
        Example 3: {\newline}
        Input: left = 1, right = 2147483647{\newline}
        Output: 0{\newline}
        
        Hint: {\newline}
        0 $<=$ left $<=$ right $<=$ 231 - 1{\newline}
        Please use c\# to finish the problem.{\newline}

        \textbf{Response:}
        {\newline}
        \texttt{The following are reference implementation options:{\newline}
            ```cs{\newline}
            public class Solution \{{\newline}
            \hspace*{2em}public int RangeBitwiseAnd(int left, int right) 
            \{{\newline}
            \hspace*{2em}\hspace*{2em}while (left < right) \{{\newline}
            \hspace*{2em}\hspace*{2em}\hspace*{2em}right \&= (right - 1);{\newline}
            \hspace*{2em}\hspace*{2em}\}{\newline}
            \hspace*{2em}\hspace*{2em}return right;{\newline}
            \hspace*{2em}\}{\newline}
            \}{\newline}
            ```{\newline}
        }
        \end{minipage}
\\ \midrule
\end{longtable*}

\begin{figure}[h]
    \centering
    \footnotesize
    \begin{tabular}{p{0.3\linewidth} p{0.3\linewidth}p{0.3\linewidth}}
    \toprule
    \multicolumn{3}{l}{\textbf{Prompt:} }  \\
    \multicolumn{3}{p{0.95\linewidth}}{
    A: A young music enthusiast who is passionate about listening to music by actors and female singers in the pop genre.

    \vspace{4pt}
    
    B: \textbf{An older music enthusiast who enjoys listening to classical and jazz music}.

    \vspace{4pt}
    
    Generate a snippet of their conversation.
    \vspace{5pt}
    } \\
    \midrule
    \textbf{Ground Truth:} & \textbf{Fine-grained corruption:} & 
    \textbf{Coarse-grained corruption:} \\
    
    \vspace{4pt}
    
    [history conversations]

    \vspace{4pt}
    
    A: I see. Do you have any other music recommendations?
    
    \vspace{4pt}
    
    B: I've been listening to \textbf{Billie Holiday's jazz music} lately. Are you interested in giving it a try?
    \vspace{2pt}
    &
    
    \vspace{4pt}
    
    [history conversations]

    \vspace{4pt}
    
    A: I see. Do you have any other music recommendations?
    
    \vspace{4pt}
    
    B: I've been listening to \textbf{Justin Bieber's jazz music} lately. Are you interested in giving it a try?
    \vspace{2pt}
    &
    
    \vspace{4pt}
    
    A: Hello, I'm A. Lately, I've gotten into cooking and I especially enjoy trying out various new recipes. Are you interested in cooking?

    \vspace{4pt}
    
    B: Hello, I'm B. Actually, I'm quite interested in cooking too, especially baking. Recently, I've learned to make a delicious chocolate cake.

    \vspace{4pt}
    [folded conversations]
    \vspace{2pt}
        \\
    \midrule[\heavyrulewidth]

    \multicolumn{3}{l}{\textbf{Prompt:} }  \\
    \multicolumn{3}{p{0.95\linewidth}}{

    [...]
    
    According to foreign media reports, Microsoft announced that it would lay off 10,000 employees, about \textbf{5\%} of its total workforce, by the end of March. On January 20th, Google suddenly announced a layoff of 12,000 people, accounting for \textbf{6\%} of its workforce. On January 26th, The Wall Street Journal reported that IBM had joined the ranks of layoffs and would cut about 3,900 people...
    
    [...]
    
    Write a news article based on the given information.
    \vspace{5pt}
    } \\
    \midrule
    \textbf{Ground Truth:} & \textbf{Fine-grained corruption:} & 
    \textbf{Coarse-grained corruption:} \\
    \vspace{1pt}
    Title: After the Chinese New Year Holiday, Wave of Layoffs Hits - Cherishing Work Becomes the Greatest Sense of Security for Adults

    \vspace{4pt}
    Article:
    
    [...]

    \vspace{4pt}
    According to foreign media reports, Microsoft has announced plans to lay off 10,000 employees by the end of March, accounting for approximately \textbf{5\%} of its total workforce. On January 20th, Google made a sudden announcement to lay off 12,000 people, representing \textbf{6\%} of its total workforce. On January 26th, The Wall Street Journal reported that IBM has also joined the layoff trend, planning to cut approximately 3,900 jobs.
    
    \vspace{4pt}
    [...]
    \vspace{1pt}
    &

   \vspace{1pt}
    Title: The Chinese New Year Holiday Ends, Most Companies Have Ceased Operations

    \vspace{4pt}
    Article:

    [...]

    \vspace{4pt}
    According to foreign media reports, Microsoft plans to lay off 1,000 employees by the end of March, accounting for approximately \textbf{10\%} of its total workforce. On January 20th, Google made a sudden announcement to lay off 12,000 people, representing \textbf{60\%} of its total workforce. The Wall Street Journal reported on January 26th that IBM has also joined the wave of layoffs, intending to terminate approximately 3,900 employees.
    
    \vspace{4pt}
    [...]
    \vspace{1pt}
    &
    \vspace{1pt}
    According to the latest reports, several cities in China have recently experienced snow disasters and cold wave weather, causing significant disruptions to people's lives and travel. It has been reported that heavy snowfall has led to the closure of many major transportation routes, resulting in a large number of vehicles stranded on the roads and many people trapped in traffic jams. Additionally, power outages in many cities due to the snow disaster have inconvenienced residents' daily lives.
    \vspace{1pt}
        \\
    \bottomrule
    \end{tabular}
    \caption{Examples of distractors derived from manually corrupted Role-play Chat and Creative-writing output.}
    \label{fig:creative_writing_fine_course}
\end{figure}

\begin{figure}[tp]
    \centering
    \footnotesize
    \begin{tabular}{p{0.48\linewidth} p{0.48\linewidth}}
    \toprule

    \multicolumn{2}{l}{\textbf{Prompt:} }  \\
    \multicolumn{2}{p{0.95\linewidth}}{
        \begin{minipage}{\linewidth}\vspace{5pt}
        Given a binary array "nums", you need to remove one element from it. Return the length of the longest non-empty subarray containing only "1" in the resulting array. If there is no such subarray, return "0". [folded examples] Please use python3 to finish it.
        \end{minipage}\vspace{5pt}
    } \\
    \midrule
    \textbf{Ground Truth:} \\
        \begin{minipage}{2\linewidth}
        \vspace{5pt}\texttt{The following is a reference implementation for your consideration:\\
            ```python\\
            class Solution:\\
            \hspace*{2em} def longestSubarray(self, nums: List[int]) -> int:\\
            \hspace*{2em} \hspace*{2em} n = len(nums)\\
            \hspace*{2em} \hspace*{2em} left = [0] * n\\
            \hspace*{2em} \hspace*{2em} right = [0] * n\\
            \hspace*{2em} \hspace*{2em} for i in range(1, n):\\
            \hspace*{2em} \hspace*{2em} \hspace*{2em} if nums[i - 1] == 1:\\
            \hspace*{2em} \hspace*{2em} \hspace*{2em} \hspace*{2em} left[i] = left[i - 1] + 1\\
            \hspace*{2em} \hspace*{2em} for i in \textbf{range(n - 2, -1, -1)}:\\
            \hspace*{2em} \hspace*{2em} \hspace*{2em} if nums[i + 1] == 1:\\
            \hspace*{2em} \hspace*{2em} \hspace*{2em} \hspace*{2em} right[i] = right[i + 1] + 1\\
            \hspace*{2em} \hspace*{2em} return max(a + b for a, b in zip(left, right))\\
            ```
        }
        \end{minipage}\vspace{5pt}
    \\
    \textbf{Fine-grained corruption:} \\
        \begin{minipage}{2\linewidth}
        \vspace{5pt}\texttt{The following is a reference implementation for your consideration:\\
            ```python\\
            class Solution:\\
            \hspace*{2em} def longestSubarray(self, nums: List[int]) -> int:\\
            \hspace*{2em} \hspace*{2em} n = len(nums)\\
            \hspace*{2em} \hspace*{2em} left = [0] * n\\
            \hspace*{2em} \hspace*{2em} right = [0] * n\\
            \hspace*{2em} \hspace*{2em} for i in range(1, n):\\
            \hspace*{2em} \hspace*{2em} \hspace*{2em} if nums[i - 1] == 1:\\
            \hspace*{2em} \hspace*{2em} \hspace*{2em} \hspace*{2em} left[i] = left[i - 1] + 1\\
            \hspace*{2em} \hspace*{2em} for i in \textbf{range(-1, n-2, 1)}:\\
            \hspace*{2em} \hspace*{2em} \hspace*{2em} if nums[i + 1] == 1:\\
            \hspace*{2em} \hspace*{2em} \hspace*{2em} \hspace*{2em} right[i] = right[i + 1] + 1\\
            \hspace*{2em} \hspace*{2em} return max(a + b for a, b in zip(left, right))\\
            ```
        }
        \end{minipage}\vspace{5pt}
    \\
    \textbf{Coarse-grained corruption:} \\
        \begin{minipage}{2\linewidth}
        \vspace{5pt}\texttt{The following is a reference implementation for your consideration:\\
            ```python\\
            def find\_max\_sum(nums):\\
            \hspace*{2em} max\_sum = float('-inf') \\
            \hspace*{2em} curr\_sum = 0 \\
            \hspace*{2em} for num in nums: \\
            \hspace*{2em} \hspace*{2em} curr\_sum += num \\
            \hspace*{2em} \hspace*{2em} if curr\_sum > max\_sum:\\
            \hspace*{2em} \hspace*{2em} \hspace*{2em} max\_sum = curr\_sum\\
            \hspace*{2em} \hspace*{2em} if curr\_sum < 0:\\
            \hspace*{2em} \hspace*{2em} \hspace*{2em} curr\_sum = 0\\
            \hspace*{2em} return max\_sum\\
            \\
            def find\_longest\_subarray(nums):\\
            \hspace*{2em} n = len(nums)\\
            \hspace*{2em} max\_len = 0\\
            \hspace*{2em} for i in range(n):\\
            \hspace*{2em} \hspace*{2em} for j in range(i, n):\\
            \hspace*{2em} \hspace*{2em} \hspace*{2em} subarray = nums[i:j+1]\\
            \hspace*{2em} \hspace*{2em} \hspace*{2em} if find\_max\_sum(subarray) == len(subarray):\\
            \hspace*{2em} \hspace*{2em} \hspace*{2em} \hspace*{2em} max\_len = max(max\_len, len(subarray))\\
            \hspace*{2em} return max\_len\\
            ```
        }
        \end{minipage}\vspace{5pt}
    \\
    \bottomrule
    \end{tabular}
    \caption{Examples of distractors derived from manually corrupted Code Generation output.}
    \label{fig:code_fine_course}
\end{figure}

\begin{figure*}[]
  \centering
  \includegraphics[width=1\linewidth]{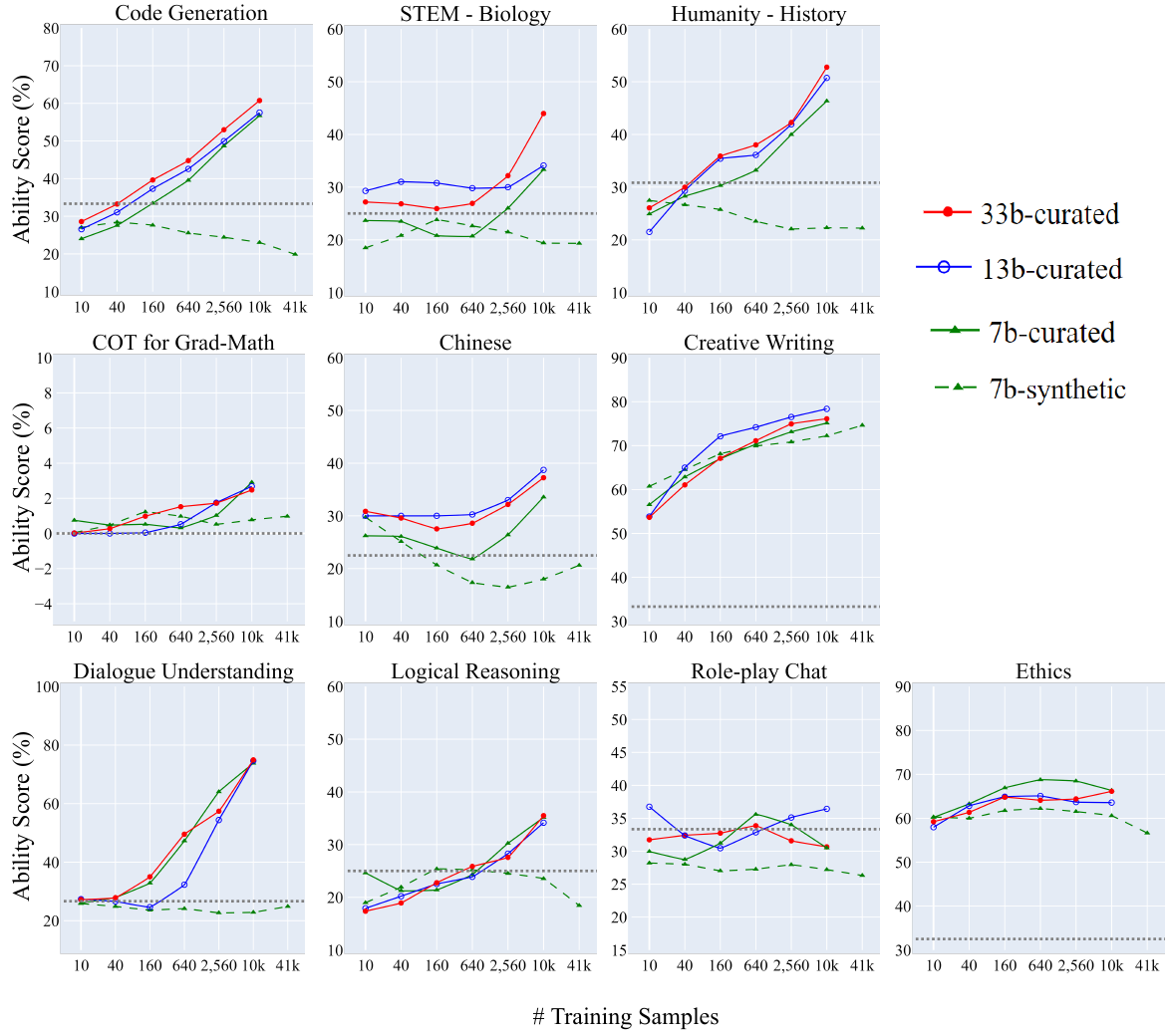}

  \caption{Evaluations of 7b models trained on synthetic data, yielding analogous conclusions as 13b models.}

  \label{fig:7b}

\end{figure*}

\newpage
\subsection{Mix up with Syhtnetic Data}
\label{mix}

\begin{table}[t]
% \vspace{-11mm}
\centering
\Huge
\resizebox{0.7\linewidth}{!}{%
\begin{tabular}{@{}lccccclll@{}}
\toprule
\multirow{2}{*}{Models} & \multirow{2}{*}{Quantity} & \multicolumn{3}{c}{AGIEval - 0shot}              & \multirow{2}{*}{} & \multicolumn{3}{c}{CMMLU - 0shot}       \\ \cmidrule(lr){3-5} \cmidrule(l){7-9} 
                &             & ep05           & ep10           & ep15           &  & \multicolumn{1}{c}{ep05} & \multicolumn{1}{c}{ep10} & \multicolumn{1}{c}{ep15} \\ \midrule
Maximum$+$0     & 40k$+$0     & 36.41          & 37.61          & 38.85          &  & \textbf{37.74}           & 37.28                    & 37.53                    \\
Maximum$+$2.56k         & 40k$+$2.56k                    & \textbf{37.08} & \textbf{39.21} & \textbf{39.88} &                   & 37.30 & \textbf{37.74} & \textbf{37.74} \\
Maximum$+$41k   & 40k$+$41k   & 32.69          & 34.43          & 34.38          &  & 33.98                    & 36.20                    & 35.34                    \\ \midrule
\multirow{2}{*}{}       & \multirow{2}{*}{}              & \multicolumn{3}{c}{AGIEval - 5shot}              & \multirow{2}{*}{} & \multicolumn{3}{c}{CMMLU - 5shot}       \\ \cmidrule(lr){3-5} \cmidrule(l){7-9} 
                &             & ep05           & ep10           & ep15           &  & \multicolumn{1}{c}{ep05} & \multicolumn{1}{c}{ep10} & \multicolumn{1}{c}{ep15} \\ \midrule
Maximum$+$0     & 40k$+$0     & 33.37          & 33.57          & 33.35          &  & \textbf{37.02}           & \textbf{37.16}           & \textbf{37.13}           \\
Maximum$+$2.56k & 40k$+$2.56k & \textbf{34.11} & \textbf{34.07} & \textbf{34.00} &  & 36.91                    & 36.87                    & 36.46                    \\
Maximum$+$41k   & 40k$+$41k   & 30.06          & 31.65          & 31.41          &  & 34.07                    & 35.06                    & 35.17                    \\ \bottomrule
\end{tabular}
}
\vspace{-2mm}
\caption{Comparing the performance of three mixing strategies with synthetic data on two benchmarks, evaluated using checkpoints at epochs 5, 10, and 15 with a parameter size of 7b. Highest performance under each setting is in bold.}
\label{tab:mixup}
\end{table}

Synthetic data is a rich open resource, but Section \ref{sec:analysis} indicates that "\textit{Synthetic data does not consistently enhance model performance with increased volume.}" Investigating the optimal use of synthetic data alongside human-curated data is crucial for practical applications. We thus utilize the "Maximum" construction from Section \ref{recon} as our baseline and then integrate varying quantities (0, 2560, 41k) of synthetic data to train 7b models.

Table \ref{tab:mixup} presents the efficacy of three mixing strategies at epochs 5, 10, and 15, evaluated on two benchmarks in both 0-shot and 5-shot scenarios. For AGIEval, a modest addition of 2,560 synthetic instances shows performance gains. In contrast, for CMMLU, peak performance is achieved with no synthetic data or a similar modest addition. Notably, further incorporating 41k synthetic instances does not yield further benefits and instead results in a marked decrease in performance, consistent with our earlier observations in Section \ref{sec:study}.

\end{document}